\documentclass[12pt,onecolumn,a4paper]{article}

\usepackage{cvpr}              

\usepackage{titlesec}
\usepackage[style=apa, backend=biber]{biblatex}
\usepackage{fancybox}
\usepackage{indentfirst}
\titleformat{\section}{\normalsize\bfseries}{\thesection}{1em}{}
\titleformat{\subsection}{\normalsize\bfseries}{\thesubsection}{1em}{}
\titleformat{\subsubsection}{\normalsize\bfseries}{\thesubsubsection}{1em}{}

\usepackage[top=2.5cm, bottom=2.5cm, left=2.5cm, right=2.5cm]{geometry}
\usepackage{setspace}
\DeclareFieldFormat{apacase}{#1}
\definecolor{cvprblue}{rgb}{0.21,0.49,0.74}
\usepackage[breaklinks,colorlinks,allcolors=cvprblue]{hyperref}

\title{Consensus-based Agentic Large Language Model Framework for Harmonized Tariff Schedule Code Classification}

\author{Truong Thanh Hung Nguyen$^{1,*}$, Khanh Van Quynh Nguyen$^{2,*}$, Hoang-Loc Cao$^1$,\\Tri Duong$^2$, Phuc Ho$^1$, Van Pham$^1$, Loc Nguyen$^1$, Hung Cao$^1$\\
$^1$Analytics Everywhere Lab, University of New Brunswick, Canada\\
$^2$University of Economics Ho Chi Minh City, Vietnam\\
{\tt\small hung.ntt@unb.ca, quynhnkv@ueh.edu.vn, hung.cao@unb.ca}
}

\begin{document}
\maketitle
\begin{abstract}
Accurate Harmonized Tariff Schedule (HTS) code classification is essential for customs clearance, duty assessment, trade statistics, and regulatory compliance in maritime logistics. However, exact HTS classification remains challenging because product descriptions are often short, incomplete, or ambiguous, while correct classification depends on hierarchical tariff structures, legal notes, and jurisdiction-specific rules. This paper proposes an agentic large language model (LLM) framework for Canadian 10-digit HTS code classification in smart-port and maritime logistics environments. The framework integrates multi-agent information retrieval, semantic retrieval over official tariff documents, evidence-grounded reasoning, consensus-based validation, element-wise voting across hierarchical code components, confidence estimation, and human-in-the-loop escalation. We evaluate the framework on a private dataset of 3,300 domain-expert-labeled product records collected from logistics and delivery contexts. Experimental results show that exact 10-digit classification remains difficult even for advanced LLMs, with performance decreasing from coarse chapter-level prediction to fine-grained tariff and statistical suffix assignment. These findings demonstrate the need for evidence-grounded, uncertainty-aware, and human-centered classification workflows rather than fully autonomous single-step prediction. The proposed framework supports more interpretable, accountable, and compliance-oriented HTS classification for maritime logistics and smart-port operations. Our code is available at \url{https://github.com/Analytics-Everywhere-Lab/hts}.

\end{abstract}

\section{Introduction}
The digital transformation of maritime logistics has accelerated the development of smart ports and intelligent supply chain systems \parencite{LiuYuen2025AISeaports, UNCTAD2024RMT}.  These systems rely on data-driven decision making, automation, AI, and human-centered platforms to improve efficiency, reduce clearance delays, and strengthen global trade resilience \parencite{TeixeiraEtAl2026SmartGates, DagistanEtAl2025SmartPorts}. Within this context, customs documentation and regulatory compliance remain critical bottlenecks \parencite{yuvraj2025atlas, UNCTAD2024RMT}. Among them, Harmonized Tariff Schedule (HTS) code classification is especially important because it directly affects customs clearance, duty assessment, trade statistics, preferential tariff treatment, and regulatory compliance \parencite{CBSAGuideOrigins, ITA2025HSCodes}.

The Harmonized System (HS), developed by the World Customs Organization (WCO), serves as the basis for customs tariffs and international trade statistics \parencite{WCO2012GIR}. It provides a globally shared 6-digit structure of chapters, headings, and subheadings, while national tariff systems extend this base with jurisdiction-specific digits. For example, Viet Nam uses an 8-digit structure under the ASEAN Harmonized Tariff Nomenclature, whereas Canada requires a 10-digit classification number for tariff and statistical purposes \parencite{CBSA2026Tariff, CBSAGuideRead,VietnamMOF2022Circular31,WCOConvention1983}. Thus, HTS classification is not generic product categorization, but a jurisdiction-specific legal process governed by the GRIs, Section and Chapter Notes, national tariff provisions, and other authoritative materials.

Accurate HTS classification is both operationally and legally significant. In Canada, the assigned 10-digit code determines duty rates, supports customs accounting, and affects eligibility for preferential tariff treatment \parencite{CanadaCustomsTariffAct,CBSAGuideOrigins}. Misclassification can lead to incorrect duty payments, shipment delays, increased scrutiny, and penalties \parencite{UPS2025, CanadaCustomsAct1985s109,CBSAGuideOrigins, CBSAMasterPenaltyDocument}. This risk is amplified in maritime logistics, where ports handle large and diverse cargo flows, while commercial product descriptions are often short, incomplete, inconsistent, or compressed. Legally relevant attributes, e.g., material composition, intended use, manufacturing process, construction, or technical specifications, may be missing from the input.

Prior work has explored rule-based systems, supervised machine learning, neural networks, and large language models (LLMs) for HS/HTS classification. Rule-based and expert-assistance systems offer transparency but require extensive maintenance and are difficult to scale \parencite{grainger2024customs}. Traditional machine learning improves scalability but depends heavily on labeled data and may degrade under distribution shift \parencite{marra2026automatic}. Neural models, including recurrent networks, CNNs, transformers, and hierarchical architectures, improve semantic representation and code-level dependency modeling \parencite{ding2015auto,du2021hscodenet,he2021commodity,liao2024enhanced,zhou2022harmonized}. However, many model-centric systems still prioritize predictive accuracy while providing limited traceable legal evidence, which is problematic in auditable customs settings.

Recent LLMs create new opportunities for HTS classification because they can interpret product descriptions, infer implicit attributes, reason over candidate categories, and generate structured outputs with less task-specific preprocessing \parencite{marra2026automatic}. Domain-adapted LLMs and HTS-specific benchmarks further show potential for tariff prediction using customs rulings and expert-labeled data \parencite{yuvraj2025atlas,yang2025hscodecomp}. Nevertheless, single-step LLM classification remains risky, where models may hallucinate tariff logic, overlook legal notes, generate unsupported codes, or fail to detect insufficient product information. These limitations are especially serious because HTS classification is hierarchical and path-dependent, where an error at the chapter or heading level can invalidate all lower-level digits.

This gap motivates the use of agentic LLMs for HTS code classification. In this paper, we propose an agentic LLM framework for Canadian 10-digit HTS classification in maritime logistics and smart-port environments. The framework treats HTS classification as an evidence-based, structured prediction task and integrates multi-agent retrieval, semantic search over official tariff documents, consensus-based validation, confidence estimation, and human-centered escalation to produce interpretable and accountable outputs.

The main contributions of this paper are summarized as follows:

\begin{enumerate}
    \item \textbf{We introduce a consensus-based agentic LLM framework for Canadian 10-digit HTS classification} that integrates web-based product evidence gathering and retrieval-augmented semantic search over authoritative tariff documents. This design supports evidence-grounded classification in smart-port and maritime logistics settings.

    \item \textbf{We propose a consensus-based validation mechanism for hierarchical HTS outputs.} The mechanism combines perturbation-based self-consistency, element-wise majority voting, and level-wise confidence estimation to quantify uncertainty across chapter, heading, subheading, tariff item, and statistical suffix components.

    \item \textbf{We develop a human-centered escalation workflow} that activates when classification confidence is low. The system generates targeted clarification questions about missing product attributes, enabling human users to resolve ambiguity while preserving the efficiency of automated classification.

    \item \textbf{We evaluate the proposed framework on a private dataset of 3,300 domain-expert-labeled Canadian HTS records} collected from logistics and delivery contexts. The evaluation reports both level-wise and whole-code accuracy across multiple LLMs, demonstrating the difficulty of exact 10-digit classification and the importance of hierarchical, uncertainty-aware evaluation.
\end{enumerate}
\section{HTS Code Classification: Background and Related Work}


Adopted by more than 200 countries and economies, the HTS classification is a massive legal instrument of customs tariffs and international trade, organizing thousands of commodity groups across 99 chapters \parencite{yuvraj2025atlas,WCO2012GIR}. This structure is consistently implemented across major trading nations, reflecting the universal adoption of the WCO-developed HS. For instance, Canada organizes its customs tariff schedule into 21 sections and 99 chapters \parencite{CBSA2026Tariff, CBSAGuideCCT}, the United States (US) has 22 sections and 99 chapters \parencite{USITC2024HTS}, and Vietnam operates under an equivalent framework within ASEAN guidelines \parencite{VietnamMOF2022Circular31}. Notably, the HTS undergoes amendments every five years to reflect changes in global trade policies, technological advancements, and new international agreements \parencite{WCO2025HSMultiPurpose}. The latest version of the HS came into effect on 1 January 2022, and the upcoming amendment is scheduled for 2028 \parencite{WCO2025HSMultiPurpose}. This section examines two foundational dimensions of the HTS: the hierarchical structure of the code itself and the legal process through which the accurate classification of traded products is determined. 

\subsection{HTS Code Structure}
The structure of the HTS code consists of two tiers: a globally shared 6-digit HS base followed by country-specific national extensions of 2 to 4 digits, forming the HTS codes unique to each country \parencite{WCOConvention1983}. The first six digits of the HTS code, corresponding to the chapter, heading, and subheading levels of the international HS, are identical for the same product across all WCO member countries and economies, creating a global common language of trade classification \parencite{CBSAGuideOrigins,WCOConvention1983}. Beyond the international 6-digit HS base, the HTS codes diverge across jurisdictions through country-specific national extensions \parencite{WCOConvention1983}. For example, in Canada, digits 7-8 represent the first Canadian subheading, forming the ``rate line'' and allowing the \textcite{CBSA2026Tariff} to determine the duty payable on imported goods \parencite{UPS2025,CBSAGuideRead}. Digits 9-10 constitute the statistical suffix used for trade data collection \parencite{CBSAGuideRead}. These last four national digits are legally valid only within the country that assigned them \parencite{WCOConvention1983,CBSAGuideRead}, meaning they vary across jurisdictions for the same product rather than being shared as the six-digit HS base. Consequently, classification must be performed separately for each importing country.

The GRI framework operates on the legal premise of single-point classification, restricting every product traded in the global market to exactly one correct classification point within the HS nomenclature \parencite{WCO2012GIR, CBSAGuideGRI}. To identify this single classification point, six GRIs codify the legal principles that must be applied sequentially in every classification of any product in order to determine its correct tariff position \parencite{WCO2012GIR, CBSAGuideGRI}. This means that each subsequent rule is invoked only when the preceding rule is insufficient to determine the product's classification \parencite{WCO2012GIR, CBSAGuideGRI}. Specifically, GRI 1 determines the classification of goods based on the text of the headings and legally binding Section and Chapter Notes, resolving the vast majority of products at this first step \parencite{WCO2012GIR, CBSAGuideGRI}. Subsequently, GRI 2 broadens the scope of the headings by capturing incomplete or unfinished articles and mixtures or combinations of multiple materials and substances \parencite{WCO2012GIR, CBSAGuideGRI}. GRI 3 addresses cases where a product is prima facie classifiable under two or more headings, providing the priority order in applying the following sub-rules: the most specific description (GRI 3a), the material or component that gives the item its essential character (GRI 3b), and the heading that occurs last in numerical order (GRI 3c) \parencite{WCO2012GIR, CBSAGuideGRI}. GRI 4 serves as a residual rule for unclassifiable goods under GRIs 1-3, directing the classifier to the heading that covers the most akin goods \parencite{WCO2012GIR, CBSAGuideGRI}. GRI 5 governs containers and packing materials by generally classifying them with the goods they accompany \parencite{WCO2012GIR, CBSAGuideGRI}. GRI 6 extends the application of GRIs 1-5 to the subheading level, requiring comparison only among subheadings at the same level \parencite{WCO2012GIR, CBSAGuideGRI}. In Canada, the GRIs are further supplemented by three Canadian rules, which are applied after GRI 6 to support Canadian domestic classification at the tariff item level \parencite{CBSAGuideCanadianRules}. 

Despite this structured legal framework, any misclassification could create significant legal and financial consequences. In terms of financial consequences, the assigned HTS code is critical for determining the applicable duty rate for imported goods, eligibility for preferential tariff treatment under free trade agreements (FTAs), and compliance with non-tariff measures such as import licensing and trade remedy orders \parencite{UPS2025}. As to the legal consequences, misclassification of traded products constitutes a legal violation under the Canadian Customs Act \parencite{CanadaCustomsAct1985s109}. Additionally, importers have a legal obligation to classify goods correctly rather than the customs authorities \parencite{CanadaCustomsAct1985s32}. Furthermore, misclassification is also subject to the AMPS under the Customs Act, with penalties escalating for repeated findings of non-compliance \parencite{CBSAMasterPenaltyDocument, CanadaCustomsAct1985s109}. Consequences of violating the Customs Act include shipment delays, increased customs scrutiny, potential seizure of goods, and reputational damage \parencite{CBSAGuideOrigins,UPS2025}. At a global scale, rapid changes in tariff classification requirements can have significant operational consequences, as exemplified by the 2025 changes to US tariff policies \parencite{UPU2025DeMinimis}. According to the \textcite{UPU2025DeMinimis}, 25 of its 192 members suspended outbound postal services to the US due to concerns over operational readiness and regulatory alignment surrounding new customs duty collection requirements. These operational demands are further amplified in the maritime port settings, given this sector’s narrow clearance windows, high cargo volumes, and diverse product descriptions. These conditions create significant classification demands that manual processes alone cannot meet. This underscores the need for automated, evidence-based classification support within customs and port operations, which the intelligent classification framework examined in this paper addresses. 

\subsection{HTS Code Classification Task and Computational Techniques}

\paragraph{Task Definition}
HTS code classification is the process of assigning a traded product to its legally correct tariff code under the HS or a country-specific HTS. Given an input product description, the objective is to infer a structured code path through the tariff hierarchy, from broad product categories such as chapters and headings to more fine-grained national tariff items and statistical suffixes. Unlike ordinary text classification, HTS classification is a legally constrained hierarchical prediction task: the correctness of lower-level digits depends on the correctness of higher-level categories, and the final output must be consistent with the General Rules for the Interpretation of the Harmonized System, Section and Chapter Notes, national tariff provisions, and other authoritative classification materials. This makes the task especially challenging in real logistics settings, where product descriptions are often short, incomplete, noisy, or ambiguous, and where important classification attributes such as material composition, intended use, construction, or technical specifications may be missing \parencite{du2021hscodenet,lee2024explainable,marra2026automatic}.

\paragraph{Prior Approaches}
Early approaches to HTS and HS classification relied heavily on rule-based search, expert systems, and manual decision-support tools. These systems typically assist human classifiers by providing access to tariff schedules, explanatory notes, customs rulings, and keyword-based search over product categories. Their main advantage is legal transparency, which allows classification decisions to be traced to explicit tariff headings, notes, rulings, and interpretive rules. Official customs guidance emphasizes that tariff classification is not merely a keyword-matching exercise but a legal reasoning process grounded in the HS text, legal notes, explanatory notes, and classification opinions. Public systems, e.g., customs ruling databases and product-code lookup tools, reflect this search-and-support paradigm, where technology helps users locate relevant legal materials but does not fully automate final classification. However, rule-based and expert-system approaches are difficult to scale because they require extensive expert knowledge engineering, continuous maintenance as tariff schedules change, and careful handling of exceptions. They are also limited when product descriptions are vague or when the relevant classification depends on implicit product attributes not stated in the input. Hence, such systems are useful for expert assistance but insufficient as fully automated classifiers in high-volume logistics environments \parencite{grainger2024customs}.

To improve scalability, later studies formulated HS/HTS classification as a supervised learning problem. Traditional models have been applied to customs declarations or product descriptions after substantial preprocessing, including tokenization, normalization, stop-word removal, lemmatization, and feature engineering \parencite{marra2026automatic}. These methods can be effective when training and test data come from similar custom domains, but conventional ML models are often brittle under distribution shift, where their performance can degrade substantially on external custom domains and product-description datasets. Moreover, conventional classifiers often treat HTS prediction as a flat-label classification problem, even though the tariff code itself is hierarchical and legally path-dependent, which limits their ability to model dependencies between HTS code components, also noted in the broader hierarchical classification literature \parencite{silla2011survey}.

Neural network (NN) methods address some of these limitations by learning richer semantic representations from product text. Recurrent neural networks (RNNs), convolutional neural networks (CNNs), transformer-based encoders, and hybrid architectures have been explored for commodity and HS code classification \parencite{ding2015auto,he2021commodity,liao2024enhanced}. A particularly relevant direction is hierarchical neural modeling, which explicitly incorporates the multi-level structure of HS codes. For example, \textcite{du2021hscodenet} propose HScodeNet, which combines hierarchical sequential information and global textual relationships to improve commodity HS code classification. Other studies use attention mechanisms \cite{liao2024enhanced} or hybrid CNN-transformer architectures \parencite{zhou2022harmonized} to better capture discriminative product attributes and code dependencies. These approaches are better aligned with the structure of the tariff nomenclature than flat classifiers, as they recognize that errors at higher levels propagate to lower levels. Nevertheless, neural models still require large, high-quality labeled datasets, and their predictions may remain difficult to explain in legally meaningful terms.

Another important line of work shifts from direct prediction to explainable decision support. \textcite{lee2024explainable} develop a KoELECTRA-based explainable model that recommends likely HS headings and subheadings from product descriptions. This retrieval-and-explanation framing is particularly important for HTS classification because operational usefulness depends not only on whether a model predicts a plausible code, but also on whether the decision can be inspected, justified, and audited by human experts.

More recently, LLMs have introduced a new paradigm for HTS classification. General-purpose LLMs can interpret natural language descriptions, reason over product attributes, and generate structured outputs with less task-specific preprocessing than classical ML systems. Prompt-based LLMs are attractive because they can be deployed quickly and can generalize across varied product descriptions \parencite{marra2026automatic}. Fine-tuned domain LLMs further improve performance when trained on authoritative customs rulings or expert-labeled tariff data. For example, \textcite{yuvraj2025atlas} introduce ATLAS, a benchmark and fine-tuned LLM for HTS code prediction built from U.S. customs rulings, showing that domain adaptation can improve both 6-digit and 10-digit prediction. However, single-shot LLM classification remains risky in compliance-sensitive settings. LLMs may hallucinate tariff logic, overlook legal notes, or produce plausible but unsupported classifications. Even when they predict the correct broad category, exact 10-digit classification remains difficult because the final digits often depend on narrow legal distinctions or missing product details \parencite{yuvraj2025atlas,yang2025hscodecomp}.

These limitations reveal an important gap in the existing literature. Most prior HTS classification techniques emphasize predictive accuracy, although customs classification is not only a prediction task. It is also an evidence-grounded legal reasoning task. A practical system must retrieve authoritative tariff information, reason over hierarchical dependencies, expose supporting evidence, estimate uncertainty, and identify when human clarification is required. Existing model-centric approaches often lack one or more of these capabilities.

\paragraph{Agentic LLMs for HTS Code Classification}
This gap motivates the use of agentic LLMs for HTS code classification. 
An agentic LLM refers to \textit{an LLM-based decision-support framework that autonomously or semi-autonomously coordinates planning, information retrieval, tool use, evidence-grounded reasoning, uncertainty estimation, and human-in-the-loop escalation to solve tasks} \parencite{plaat2025agentic,le2025multimedia,cao2026adaptive}.

Such a system is well-suited to HTS classification because the task requires both semantic understanding of product descriptions and legal grounding in tariff documents. An agentic framework can first gather external product evidence, retrieve relevant tariff provisions, generate candidate classifications, validate hierarchical consistency, and escalate ambiguous cases to human users. Recent agent-oriented HTS evaluations further suggest that tariff classification remains difficult even for advanced systems, reinforcing the need for retrieval quality, rule grounding, uncertainty control, and human oversight rather than unrestricted autonomous prediction \parencite{yang2025hscodecomp}.

Our work contributes to this emerging direction by proposing an agentic LLM framework for Canadian 10-digit HTS classification. The framework integrates multi-agent information retrieval, semantic retrieval over authoritative tariff documents, consensus-based validation, element-wise voting over hierarchical code components, confidence estimation, and human-centered escalation. This design addresses the main limitations of prior approaches by treating HTS classification as an evidence-grounded, uncertainty-aware, and collaborative decision-support process rather than a purely automated label prediction problem. In doing so, it supports both operational efficiency and regulatory accountability in maritime logistics and smart-port environments.
\section{Problem Formulation}
Let an item description be represented as:
$x \in \mathcal{X}$, where $\mathcal{X}$ denotes the space of unstructured natural language descriptions of goods in logistics systems (e.g., \textit{``car wash applicator''}). These descriptions are typically incomplete, ambiguous, and vary in detail, making classification challenging.
The goal is to predict a structured HTS code:
\begin{equation}
    y = (c, h, s, t, f),
\end{equation}
where each component corresponds to a hierarchical level in the tariff system:
\begin{itemize}
    \item $c$: chapter level, representing broad product categories,
    \item $h$: heading level, defining more specific categories,
    \item $s$: subheading, corresponding to the international HS code level,
    \item $t$: tariff item, which refines the classification and is typically used to determine duty rates at the national level,
    \item $f$: statistical suffix, used for country-specific reporting and regulatory purposes.
\end{itemize}

The final HTS code is constructed by concatenating these components:
$y = \text{concat}(c, h, s_1, s_2, t)$, resulting in a structured representation of the form:
\begin{equation}
  \underbrace{\text{XX}}_{\text{chapter}} \quad
\underbrace{\text{XX}}_{\text{heading}}\quad.
\underbrace{\text{XX}}_{\text{subheading}}\quad.
\underbrace{\text{XX}}_{\text{tariff item}}\quad.
\underbrace{\text{XX}}_{\text{statistical suffix}}
\end{equation}
where each pair of digits corresponds to one component in the hierarchy.
Thus, the classification task can be formulated as: $f(x) \rightarrow y$.

Due to hierarchical dependencies and the need for domain-specific reasoning, this task is described as a \textit{structured prediction problem} rather than a classification problem. Each component must be inferred not only from the input description but also in consistency with the higher-level categories.
To address this complexity, the proposed approach models $f$ as a multi-agent system that integrates information retrieval, contextual reasoning, and consensus-based validation to produce a reliable and interpretable prediction.

\section{Methodology}

\begin{figure}[h]
    \centering
    \includegraphics[width=\linewidth]{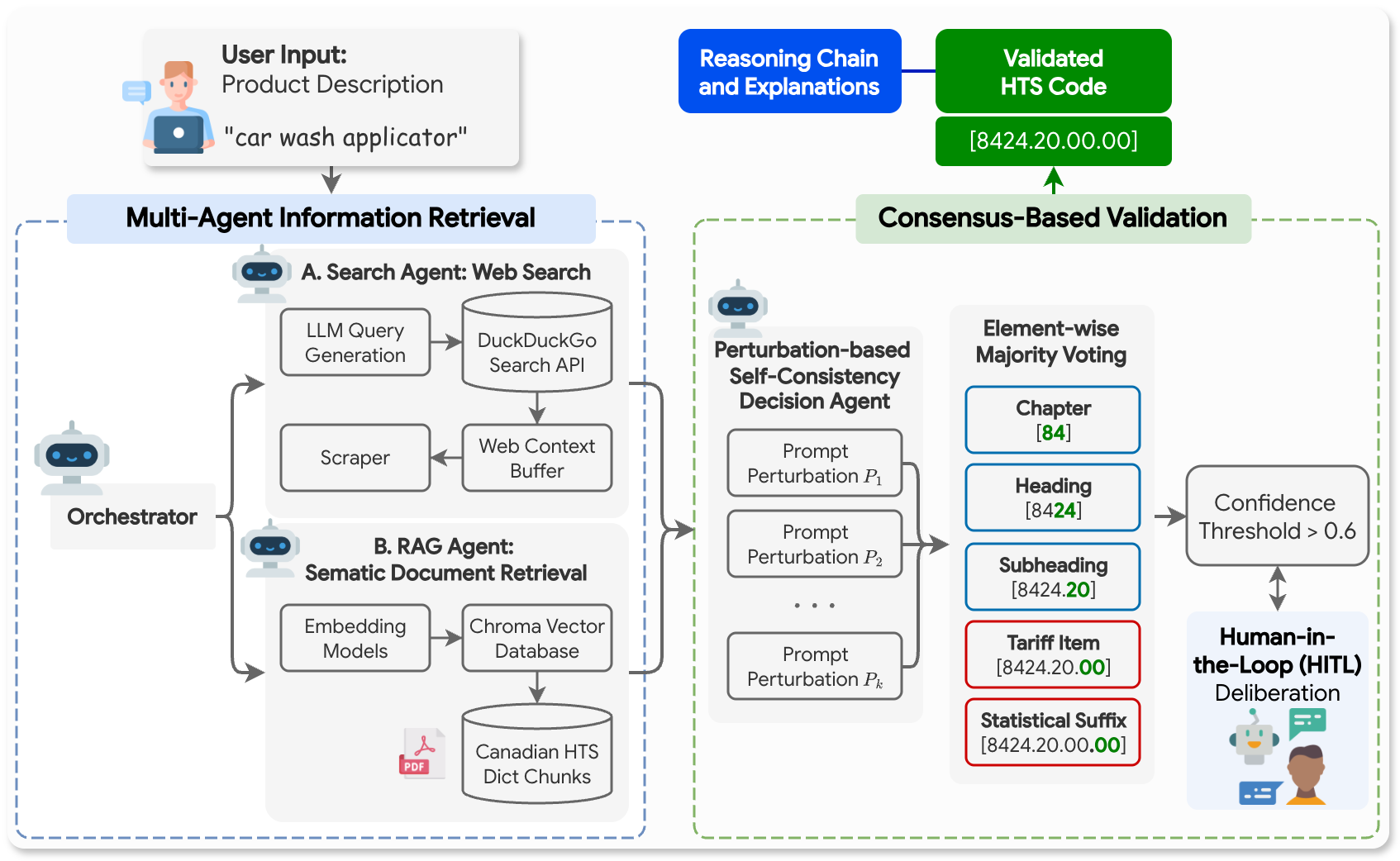}
    \caption{Architecture of the proposed HTS classification framework, showing the interaction between retrieval, reasoning, validation, and user clarification modules.}
    \label{fig:architecture}
\end{figure}

Our proposed framework is a multi-agent system designed to classify Canadian HTS codes from natural language item descriptions, as illustrated in Figure \ref{fig:architecture}.

The system is implemented as a structured workflow using a graph-based architecture, where each node represents a specific functional module.
The pipeline integrates information retrieval, reasoning, and validation to produce reliable classification results.
Our methodology structures Canadian HTS codes classification as: (i) multi-agent evidence gathering and (ii) consensus-based, uncertainty-aware decision making, and (iii) a human-centered escalation pathway. The framework is consistent with the hierarchical nature of Canadian tariff classification numbers (i.e., chapter, heading, subheading I (HS code), subheading II (determines duty), statistical suffix), where classification must be determined by applying interpretative rules and comparing candidates at the same hierarchical level. 

The overall architecture consists of two main modules: \textit{(1) Multi-Agent Information Retrieval} and \textit{(2) Consensus-Based Validation}. In addition, a human-centered interaction layer is included to handle uncertain cases.

\subsection{Module 1 – Multi-Agent Information Retrieval (MAIR)}
The Multi-Agent Information Retrieval module is designed to address a fundamental limitation in HTS classification by handling the insufficient and ambiguous nature of raw item descriptions. In real-world logistics scenarios, product descriptions are often short, incomplete, or lack critical attributes such as material composition, manufacturing process, or intended use. These attributes, however, are essential for accurate tariff classification.

To overcome this limitation, the proposed system adopts a \textit{multi-agent retrieval strategy} in which independent agents gather complementary information from heterogeneous sources. This design reflects the principle that reliable classification should be based on evidence aggregation rather than on the input text or a language model's internal knowledge alone.
Formally, the system constructs an enriched context:

\begin{equation}
    C = C_{\text{web}} \cup C_{\text{rag}},
\end{equation}
where:
\begin{itemize}
    \item $C_{\text{web}}$ represents dynamically retrieved external knowledge,
    \item $C_{\text{rag}}$ represents authoritative knowledge extracted from official tariff documents.
\end{itemize}

This dual-source design ensures that the system benefits from both adaptability and regulatory grounding.

\subsubsection{Search Agent: Web Search}
The search agent is responsible for augmenting the input description with real-world information obtained from the web. This process begins by transforming the input $x$ into a search query: $q = g_{\text{LLM}}(x),$ where $g_{\text{LLM}}$ is an LLM that generates semantically meaningful queries. Unlike naive keyword extraction, this transformation emphasizes salient attributes such as product type, material, and functional use. This step is critical because the effectiveness of downstream retrieval depends heavily on query quality.
Using the generated query, the system retrieves a set of web documents $D_{\text{web}} = \{ d_1, d_2, ..., d_k \}$. These documents are then processed through a content extraction pipeline: 
\begin{equation}
    C_{\text{web}} = \bigcup_{i=1}^{k} \text{Extract}(d_i).
\end{equation}

The extraction process filters out irrelevant HTML elements and retains meaningful textual content. In practice, this includes product descriptions, technical specifications, and contextual explanations.
From a conceptual perspective, this module enables \textit{open-domain knowledge acquisition}, allowing the system to handle novel or uncommon products, infer missing attributes, and align product descriptions with industry terminology.

This capability is particularly important in maritime logistics, where product diversity is high and standardized descriptions are not always available.

\subsubsection{RAG Agent: Semantic Document Retrieval}
While web-based information provides flexibility, it may lack legal authority and consistency. Therefore, the RAG agent retrieves information directly from the official tariff schedule, ensuring that predictions are grounded in regulatory definitions.
The tariff document is first preprocessed into a collection of text segments $\mathcal{D}_{\text{tariff}} = \{ t_1, t_2, ..., t_n \}$. 

Each segment is embedded into a semantic vector space $e_i = \phi(t_i)$, where $\phi$ is an embedding function that captures semantic meaning.
Given the input description ( x ), its embedding is computed: $e_x = \phi(x)$.
Relevant document segments are retrieved using similarity search as:
\begin{equation}
    D_{\text{rag}} = \text{Top}^{k}(\text{sim}(e_x, e_i)), \quad     C_{\text{rag}} = \bigcup_{t \in D_{\text{rag}}}.
\end{equation}

This mechanism ensures that the system has access to formal definitions of tariff categories, inclusion/exclusion rules, and hierarchical relationships between classifications.
Unlike traditional rule-based systems, this approach allows flexible with grounded reasoning, where the LLM interprets retrieved content in relation to the input.

Overall, the MAIR module provides a robust foundation for HTS classification by transforming incomplete input descriptions into rich, evidence-supported representations. This enables more accurate, interpretable, and reliable predictions in subsequent stages of the pipeline.

\subsection{Module 2 – Consensus-Based Validation}
While the MAIR module enriches the input with relevant knowledge, it does not guarantee that the final prediction will be reliable. LLMs may produce inconsistent or unstable outputs, especially when dealing with ambiguous or incomplete inputs. This issue is particularly critical in HTS classification, where incorrect predictions can lead to regulatory and financial consequences.

To address this challenge, the proposed system introduces a consensus-based validation module, which transforms the prediction process from a single deterministic output into a structured decision-making procedure. This module integrates multiple reasoning attempts, aggregates their outputs, and evaluates confidence at a fine-grained level.

\subsubsection{Perturbation-based Self-Consistency Decision Agent}
The first step in the validation process is to generate multiple predictions using prompt perturbation. Each prediction is defined as:
$y^{(i)} = f_{\text{LLM}}(x, C, p_i),$
where $x$ is the input description, $C$ is the fused context from the MAIR module, $p_i$ is a perturbed version of the prompt. The perturbations are designed to introduce slight variations in phrasing, instruction emphasis, or reasoning guidance. Although these changes are minor, they can lead the LLM to explore different reasoning paths.

From a conceptual perspective, this process serves two purposes: (1) Exploration of reasoning space: Each perturbation allows the model to interpret the same problem from a slightly different perspective, reducing the risk of systematic bias;
(2) Stability assessment: If multiple independent runs produce similar outputs, this indicates that the prediction is stable and less sensitive to prompt variation.
This mechanism can be viewed as a form of \textit{stochastic reasoning approximation}, where the system samples from the model’s implicit reasoning distribution.

\paragraph{Element-wise Majority Voting}
Each prediction is decomposed into its hierarchical components: $y^{(i)} = (c^{(i)}, h^{(i)}, s_1^{(i)}, s_2^{(i)}, t^{(i)}).$
Instead of treating the HTS code as a single label, the system performs voting at the component level. For each element $e \in {c, h, s_1, s_2, t}$, the final value is determined by:
\begin{equation}
    \hat{e} = \arg\max_{v} \sum_{i=1}^{N} \mathbb{I}(e^{(i)} = v).
\end{equation}

This element-wise approach is particularly important due to the hierarchical nature of HTS codes. In many cases, higher-level components, such as chapters and headings, are easier to predict, while lower-level components, such as tariff items or statistical suffixes, are more ambiguous. By decomposing the prediction, the system achieves error isolation, so incorrect predictions in one component do not affect others; hierarchical consistency, since each level can be validated independently in a way that reflects how human experts approach classification; and improved robustness, because aggregating multiple predictions reduces the impact of outliers.
The final HTS code is then reconstructed as:
\begin{equation}
    \hat{y} = (\hat{c}, \hat{h}, \hat{s}_1, \hat{s}_2, \hat{t}).    
\end{equation}

\paragraph{Confidence Estimation}
To quantify the reliability of each component, the system computes a confidence score:
\begin{equation}
    \text{Conf}(e) = \frac{\max_{v} \text{count}(v)}{N}.
\end{equation}

This score represents the proportion of predictions that agree on the selected value.
The use of element-wise confidence provides several advantages: it enables fine-grained uncertainty detection rather than relying on a single global confidence score, reflects the degree of consensus in a more interpretable way than model probabilities, and allows targeted intervention when specific components are uncertain.
A threshold $\tau$ is applied:
\begin{equation}
    \text{Conf}(e) < \tau \Rightarrow \text{uncertain}.
\end{equation}

In practice, this threshold defines the minimum level of agreement required for a component to be considered reliable.
From a decision-theoretic perspective, this step converts the system into a risk-aware classifier, where predictions are accepted only if they meet a confidence criterion.

\subsubsection{Human-in-the-Loop Deliberation}
When the system detects low confidence in any component, it initiates an interaction with the user. 
A clarification question is generated:
\begin{equation}
    q_{\text{human}} = h_{\text{LLM}}(x, C, \text{Conf}).
\end{equation}

This question is designed to target the most uncertain aspects of the classification. For example, it may ask about material composition, intended use, and manufacturing details. The user response $r$ is incorporated into the input $x' = x \cup r$. The classification process is then repeated: $\hat{y}' = f(x', C)$. This iterative refinement continues until the confidence condition is satisfied $\min_{e} \text{Conf}(e) \geq \tau$.

This mechanism reflects a collaborative intelligence framework in which human expertise complements automated reasoning. It ensures that the system remains both accurate and practical in real-world deployment.

\subsection{Human-Centered Application}

\begin{figure}
    \centering
    \shadowbox{\includegraphics[width=\linewidth]{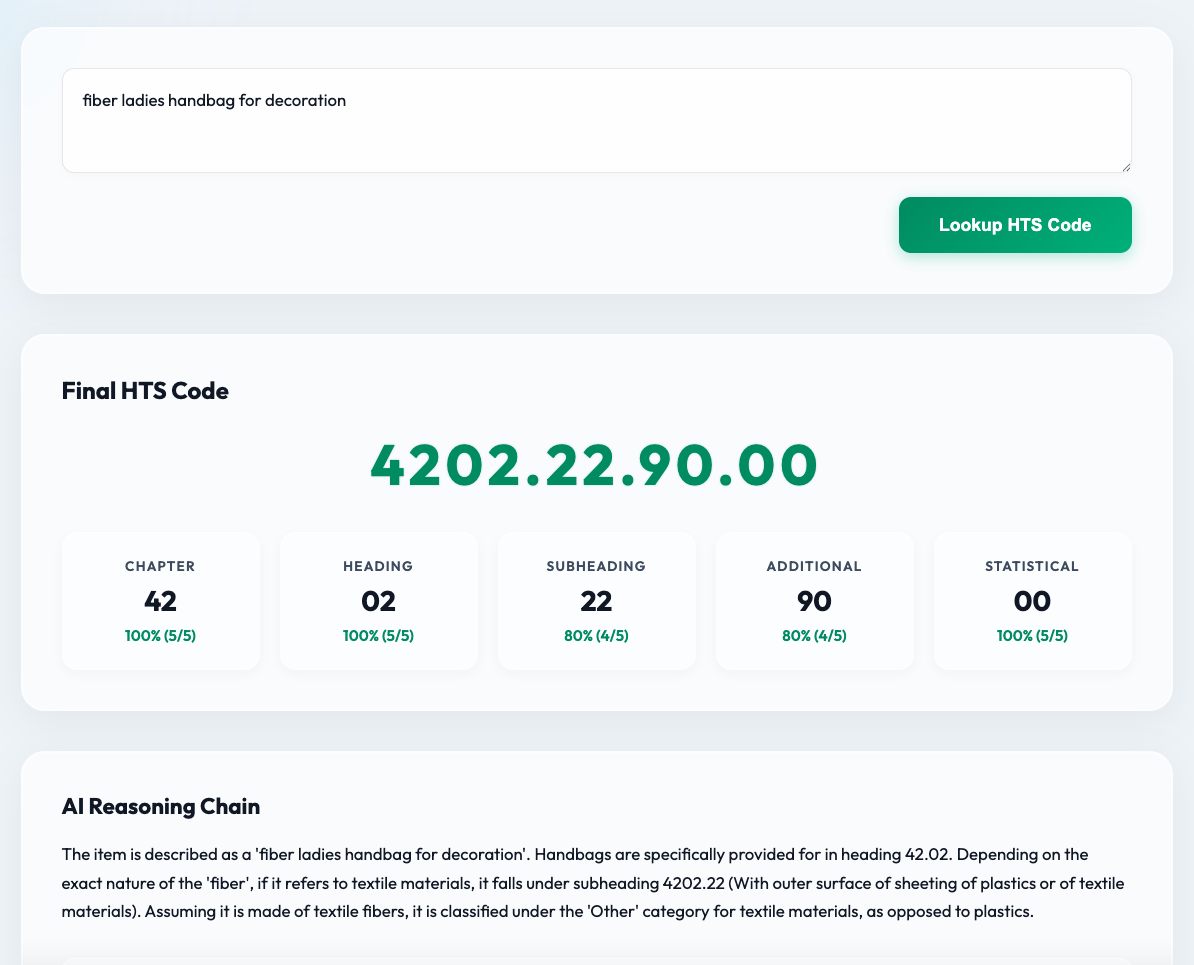}}
    \caption{Application interface of the proposed framework, showing evidence-supported HTS code prediction and transparent reasoning output.}
    \label{fig:ui}
\end{figure}

\subsubsection{HTS Code Identification with Explanations}
The HTS Code Identification with Explanations module is designed to provide an interpretation of how that classification is obtained. In the implemented system, the item description is processed together with the retrieved tariff context, and the model is instructed to return a structured output containing both the hierarchical HTS elements and a reasoning field that explains the classification logic. This design supports traceability by making the decision process visible rather than treating classification as a black-box prediction.

A key feature of this module is the decomposition of the final classification into chapter, heading, subheading, additional subheading, and statistical suffix, as illustrated in Figure~\ref{fig:ui}. This reflects the hierarchical structure of tariff classification and enables the system to present the result in a way that is easier for users to inspect and verify.

The final HTS code is reconstructed from these individual components, while the accompanying explanation summarizes the reasoning used to map the item description and retrieved evidence to the predicted code. In the application layer, these outputs are returned together with the retrieved context and reasoning steps, allowing the interface to present the classification result as an evidence-supported recommendation rather than a standalone numerical label.

This explanatory design is particularly important in customs-related decision environments, where users must often review the basis of a classification before accepting it for operational or compliance purposes. By exposing both the final code and the reasoning process, the module improves interpretability and supports more accountable use of LLM-based classification in practice.

\subsubsection{Human-AI Collaboration in Classification Workflows}

\begin{figure}
    \centering
    \shadowbox{\includegraphics[width=\linewidth]{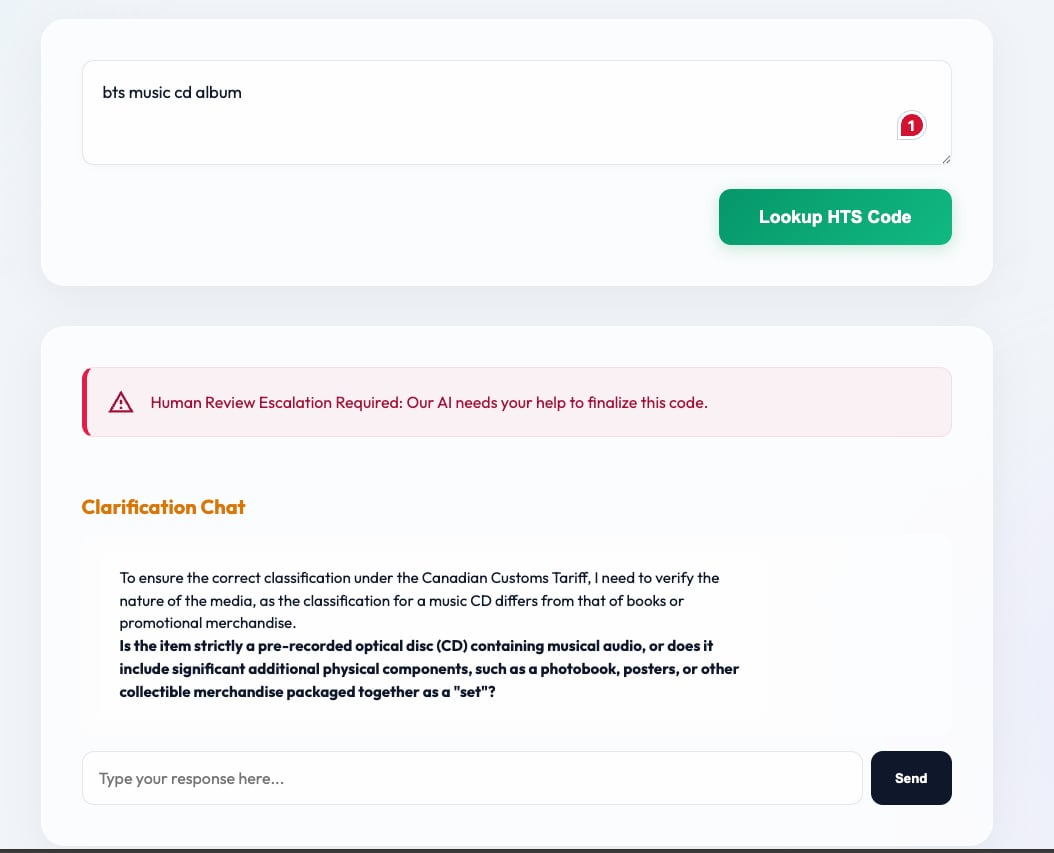}}
    \caption{Clarification chat interface activated during low-confidence HTS classification, illustrating human-AI collaboration for resolving ambiguous cases.}
    \label{fig:chat}
\end{figure}

The human-centered nature of the system is most evident in its collaborative workflow design, where automated prediction is coupled with targeted user involvement in cases of uncertainty. After generating multiple classification attempts through prompt perturbation, the system performs element-wise voting across the hierarchical HTS components and computes confidence scores for each element. This allows uncertainty to be localized to specific code segments rather than only to the full prediction. Such fine-grained confidence estimation supports a more practical interaction model, since users can understand which parts of the classification are stable and which remain ambiguous.

When the confidence of any element falls below the predefined threshold, the system activates an escalation mechanism rather than returning a potentially unreliable answer, as illustrated in Figure~\ref{fig:chat}. In this stage, the model generates a single targeted clarification question intended to elicit the most informative missing detail from the user, such as material composition or intended use.
The additional user input is then incorporated into the ongoing classification process via the escalation chat workflow, which continues until the system can produce a sufficiently confident, structured result. This interaction pattern transforms the application from a static prediction tool into a collaborative decision-support system.

From a workflow perspective, the system distributes responsibilities between AI and the human user in a complementary manner. The AI performs evidence retrieval, structured reasoning, consensus-based validation, and uncertainty detection, while the human contributes contextual clarification when the available information is insufficient for reliable classification. This design improves usability in real-world HTS classification settings, where product descriptions are often incomplete, and expert oversight remains essential. As a result, the application supports a form of human-AI collaboration that jointly optimizes efficiency, interpretability, and reliability.
\section{Experiment and Results}

\subsection{Dataset}
We evaluate the proposed framework on a private dataset collected from several delivery and logistics companies. The evaluation set contains 3,300 product records, each paired with a domain-expert-labeled 10-digit Canadian HTS code. The dataset reflects operational customs descriptions encountered in real logistics pipelines, and therefore includes both relatively specific product descriptions and highly compressed commercial shorthand. As shown in Table~\ref{tab:dataset_examples}, the product space spans multiple sectors, including apparel, footwear, industrial components, consumer goods, and sports-related merchandise. This diversity makes the task particularly challenging, since correct classification depends not only on the item's semantic category but also on subtle legal distinctions involving material composition, intended use, construction, and tariff note interpretation.

\subsection{Experimental Setup}
\paragraph{Model Configuration}
We evaluate the framework with three LLMs representing different model families and capacities: Qwen3-30B \parencite{yang2025qwen3}, GPT-OSS-120B \parencite{agarwal2025gpt}, and Gemini-3.1-Pro \parencite{GoogleDeepMind2026Gemini31}. For each item description, the system predicts a structured 10-digit HTS code composed of five hierarchical elements: chapter, heading, subheading, tariff item, and statistical suffix. This evaluation protocol follows the legal and hierarchical structure of the Canadian tariff schedule, in which lower-level digits are conditioned on the correctness of higher-level decisions.

\paragraph{Metrics} We report two types of metrics. First, we compute level-wise accuracy for each level of the HTS hierarchy. These metrics quantify how well the framework captures coarse and fine tariff structures independently. Second, we report whole-code accuracy, which requires all 10 digits to be correctly predicted. Whole-code accuracy is the strictest measure and reflects the practical requirement of exact classification for customs documentation and duty determination.

\paragraph{Experimental Setting}
To isolate the predictive capability of the automated framework itself, all experiments are conducted \textit{without activating the human-in-the-loop deliberation mechanism}. In other words, once the model retrieves contextual evidence and generates candidate classifications, the final output is produced entirely by the retrieval, reasoning, and consensus modules, without any corrective input from human users. This setting allows the evaluation to reflect the LLM-based system's intrinsic classification performance under a fully automated procedure.

\begin{table}[htbp]
\centering
\small
\caption{Representative examples from the evaluation dataset.}
\label{tab:dataset_examples}
\begin{minipage}[t]{0.48\textwidth}
\centering
\begin{tabular}{ll}
\toprule
\textbf{Description} & \textbf{HTS Code} \\
\midrule
PVC casual footwear & 6404.11.99.90 \\
Sports apparel for yoga & 6104.63.00.10 \\
Fuel filtration unit & 8421.23.00.20 \\
Latex sports balls & 9506.62.00.90 \\
Plastic switch module & 8536.50.90.90 \\
\bottomrule
\end{tabular}
\end{minipage}
\hfill
\begin{minipage}[t]{0.48\textwidth}
\centering
\begin{tabular}{ll}
\toprule
\textbf{Description} & \textbf{HTS Code} \\
\midrule
Blouse-making beads & 3926.90.99.90 \\
Plastic iron screws & 3926.90.91.20 \\
Ghost US liner shoes & 6404.11.99.21 \\
Computer case sample & 4202.92.90.00 \\
General sports goods & 9506.99.00.90 \\
\bottomrule
\end{tabular}
\end{minipage}
\end{table}

\subsection{Quantitative Evaluation}

\subsubsection{Overall Performance}

Table~\ref{tab:main_results} reports level-wise and exact-match accuracy across the HTS hierarchy. Gemini-3.1-Pro achieves the strongest overall performance, obtaining the best results at the chapter, heading, subheading, and whole-code levels. Its exact-match accuracy reaches 40.97\%, substantially outperforming GPT-OSS-120B and Qwen3-30B. This result indicates that Gemini-3.1-Pro is not only more accurate at identifying broad product categories, but also better at maintaining consistency across the dependent decisions required for full 10-digit classification.

GPT-OSS-120B ranks second. It achieves 56.25\% chapter accuracy and 46.53\% heading accuracy, showing moderate ability to recover coarse tariff categories from short product descriptions. However, its whole-code accuracy drops to 14.58\%, indicating that broad semantic understanding does not reliably transfer to fine-grained tariff assignment.

Qwen3-30B performs weakest under exact-match evaluation, with 6.94\% whole-code accuracy. Its chapter and heading accuracies are also substantially lower than those of the other models. Although Qwen3-30B obtains the highest tariff-item and suffix accuracies, these component-level gains do not translate into valid full-code predictions because HTS classification is path-dependent: lower-level predictions are only meaningful when the preceding hierarchy is correct.

\begin{table}[htbp]
\centering
\small
\caption{Level-wise and whole-code accuracy on the 3,300-item evaluation dataset (\%). The best results are in \textbf{bold}.}
\label{tab:main_results}
\begin{tabular}{l|rrrrr|r}
\toprule
\textbf{Model} & \textbf{Chapter} & \textbf{Heading} & \textbf{Subheading} & \textbf{Tariff Item} & \textbf{Suffix} & \textbf{Whole Code} \\
\midrule
Gemini-3.1-Pro & \textbf{74.31} & \textbf{61.81} & \textbf{52.08} & 61.11 & 47.92 & \textbf{40.97} \\
GPT-OSS-120B  & 56.25 & 46.53 & 42.36 & 56.94 & 43.75 & 14.58 \\
Qwen3-30B     & 23.61 & 17.36 & 18.75 & \textbf{63.89} & \textbf{48.61} & 6.94 \\
\bottomrule
\end{tabular}
\end{table}

\subsubsection{Aggregate Accuracy Across the Hierarchy}

Table~\ref{tab:summary_results} summarizes model performance using mean level-wise accuracy and whole-code exact match. Gemini-3.1-Pro again ranks first, with a mean level-wise accuracy of 59.45\%, followed by GPT-OSS-120B at 49.17\% and Qwen3-30B at 34.44\%. The agreement between the aggregate and exact-match rankings suggests that Gemini-3.1-Pro is not merely strong on a single component, but more stable across the full tariff hierarchy.

\begin{table}[htbp]
\centering
\small
\caption{Summary comparison across models. Mean level-wise accuracy (\%) is averaged over the five hierarchical HTS components. The best results are in \textbf{bold}.}
\label{tab:summary_results}
\begin{tabular}{l|rr}
\toprule
\textbf{Model} & \textbf{Mean Level-wise Accuracy ($\uparrow$)} & \textbf{Whole-Code Accuracy ($\uparrow$)} \\
\midrule
Gemini-3.1-Pro & \textbf{59.45} & \textbf{40.97} \\
GPT-OSS-120B   & 49.17 & 14.58 \\
Qwen3-30B      & 34.44 & 6.94 \\
\bottomrule
\end{tabular}
\end{table}

This comparison also highlights the limitations of relying exclusively on exact-match accuracy. Whole-code evaluation is necessary because tariff codes must be valid end-to-end, but it does not reveal where errors occur. Level-wise metrics expose whether a model fails at coarse semantic grouping, intermediate legal disambiguation, or final national reporting refinements. This distinction is important for designing downstream verification and human-in-the-loop correction strategies.

\paragraph{Hierarchy-Level Error Propagation}

A consistent pattern is that upper levels are easier than lower levels for the stronger models. Gemini-3.1-Pro declines from 74.31\% chapter accuracy to 47.92\% suffix accuracy, while GPT-OSS-120B declines from 56.25\% to 43.75\%. This degradation reflects the increasing specificity of the HTS hierarchy. Chapters and headings often correspond to broad semantic product families, whereas subheadings, tariff items, and statistical suffixes encode material composition, construction details, exceptions, and national reporting categories that may be absent from short commercial descriptions.

The gap between chapter accuracy and whole-code accuracy further quantifies error propagation. Gemini-3.1-Pro exhibits a 33.34-point gap, GPT-OSS-120B a 41.67-point gap, and Qwen3-30B a 16.67-point gap. The smaller gap for Qwen3-30B should not be interpreted as stronger hierarchical consistency, since its chapter accuracy is already low. In contrast, the larger gaps for Gemini-3.1-Pro and GPT-OSS-120B show that many residual errors arise after the broad product family has been identified. For stronger models, the main bottleneck is therefore fine-grained legal and attribute-level disambiguation rather than coarse product recognition.

\paragraph{Non-Monotonic Component Accuracy}

Performance is not strictly monotonic across the hierarchy. For both Gemini-3.1-Pro and Qwen3-30B, tariff-item accuracy exceeds subheading accuracy. This suggests that some lower-level components contain lexical cues that are easier for models to exploit than intermediate hierarchy levels. For example, product descriptions may include terms that correlate strongly with tariff-item or suffix refinements, even when the model fails to select the correct subheading.

However, such isolated component accuracy has limited practical value. HTS classification requires a coherent path from chapter to suffix, and a correct lower-level field is invalid if it is attached to an incorrect parent category. This explains why Qwen3-30B can achieve competitive tariff-item and suffix scores while still producing poor whole-code accuracy. The result underscores the importance of evaluating both component-level prediction and hierarchical validity.

\paragraph{Operational Implications}

The three models exhibit distinct deployment profiles. Gemini-3.1-Pro provides the best trade-off between level-wise robustness and exact-match correctness, making it the strongest candidate for a confidence-aware tariff classification pipeline. GPT-OSS-120B demonstrates useful coarse-level understanding but would require more frequent downstream verification to reach compliance-grade reliability. Qwen3-30B is not suitable as a standalone exact-code classifier in its current form, although its performance on selected lower-level fields suggests potential use in constrained assistive settings with strict validation.

More broadly, these results indicate that tariff-classification performance is not determined by generic model capability alone. Reliable classification requires semantic grounding, legal-rule sensitivity, retrieval quality, and preservation of hierarchical consistency across multiple dependent predictions. The best-performing system is therefore not simply the one that predicts individual components well, but the one that produces globally coherent codes under uncertainty.

\subsection{Error Analysis}
\paragraph{Ambiguity in Product Descriptions}
Qualitative inspection reveals that many errors originate from missing information in the input description rather than from purely semantic misunderstanding. A common failure mode is \emph{material ambiguity}: product descriptions often omit whether an item is made of plastic, rubber, metal, textile, leather, or composite materials, even though material composition is central to tariff assignment.

A second failure mode is \emph{functional ambiguity}. Descriptions may identify an object but omit its intended use, making it difficult to distinguish between industrial, household, medical, recreational, or other tariff-relevant categories. This is especially problematic when visually or semantically similar products fall under different headings depending on their use.

\paragraph{Missing Technical and Legal Attributes}
A third cateegory involves missing construction or specification details. Examples include whether footwear covers the ankle, whether a bag has an outer surface of a particular material, whether a garment is knitted or woven, or whether an item is a complete article rather than a part. These attributes are often decisive in tariff classification but are rarely included in short commercial descriptions.

The fourth category consists of \emph{legal-note dependencies}. Correct classification frequently requires applying exclusion rules, precedence relationships, or distinctions between neighboring headings. These cases are difficult for LLMs because the correct answer depends not only on product semantics, but also on legal structure and the interaction between multiple tariff provisions.

\paragraph{Implications for Human-in-the-Loop Classification}
These error patterns suggest that many incorrect predictions are caused by unresolved ambiguity rather than by arbitrary model behavior. This motivates the proposed human-in-the-loop module. When ensemble predictions disagree or confidence is low, the system can ask targeted clarification questions about the missing attributes most likely to affect the final code, such as material, intended use, construction, or completeness.
In this framing, the system is not merely an autonomous classifier. It functions as a decision-support tool that converts hidden ambiguity into explicit information requests. This is particularly important in customs and logistics settings, where incorrect tariff assignment can have financial, legal, and operational consequences.
\section{Discussion}

Global trade requires interoperability across jurisdictions, raising the question of whether the framework's core design principles extend beyond Canada. This section addresses that question by examining transferability to other national HTS systems and the operational implications of the framework for maritime logistics and smart ports.

\subsection{Transferability to Other HTS Code Systems}
\subsubsection{Universal Foundations Supporting Cross-Jurisdictional Transfer}
The framework is designed to be transferable to other WCO jurisdictions, as these national HTS systems share universal foundations established by WCO. As more than 200 WCO member countries universally adopt the 6-digit HS base and apply the WCO-developed GRIs, their national systems share identical chapters, headings, and subheadings, as well as classification rules with the same legal logic \parencite{WCOConvention1983}. As a result, the core reasoning logic of the framework remains unchanged in all WCO jurisdictions. 

In terms of evidence-grounded retrieval, the framework is not restricted to any specific jurisdiction. Specifically, the web search agent can retrieve product information from anywhere on the Internet, while the RAG agent only requires incorporating the target country’s official tariff documents to enrich its knowledge base. Consensus-based validation and human-in-the-loop deliberation are technical methodologies that operate on the classification output structure rather than on country-specific content. Thus, these mechanisms can function equivalently across jurisdictions. Overall, the core design principles of the proposed framework can be transferred to other jurisdictions without a fundamental redesign, requiring only targeted adaptations. 

\subsubsection{Viet Nam as a Candidate Jurisdiction and Remaining Limitations}
Viet Nam’s current HTS code system is a promising recipient of the proposed framework. Viet Nam structures its HTS code system in alignment with the WCO-developed 6-digit HS base and applies universally adopted legal rules through the six GRIs for classification \parencite{VietnamMOF2022Circular31}. Thus, this alignment minimizes the country-specific adjustments required when integrating the framework into the national system. Furthermore, Viet Nam uses an 8-digit code under the AHTN \parencite{VietnamMOF2022Circular31}, reducing the hierarchical levels the framework needs to predict compared to the Canadian 10-digit system. The last two digits of the HTS code under AHTN represent not only Viet Nam’s national extensions but also ASEAN’s regional extensions. As the AHTN framework is applicable across 10 ASEAN countries, successfully transferring the proposed classification framework to Viet Nam would also have meaningful implications for other ASEAN members, thereby extending its potential reach to the entire ASEAN region. Additionally, Viet Nam has a strong capacity for digital transformation, with its official tariff nomenclature, classification rules, and Legal Explanatory Notes established under the  Circular 31/2022/TT-BTC of the \textcite{VietnamMOF2022Circular31} (MOF) and the Official Dispatch 3866/TCHQ-TXNK of the \textcite{VietnamCustoms2023SEN} (DVC), available online, in Vietnamese and English, enabling more effective RAG retrieval. Most importantly, Viet Nam has an increasingly significant need for an automated HTS classification system given its rapid growth in maritime trade, with its Liner Shipping Connectivity Index (LSCI) recording a 199\% increase over the period from 2006 to 2024, the highest among the top 10 countries \parencite{UNCTAD2024RMT}. 

\subsubsection{Adaptation Requirements and Country-Specific Challenges}

Although the HS base is shared internationally, national tariff extensions vary substantially, limiting the direct transferability of the proposed framework. Countries differ in code length, rate-line structure, legal instruments, documentation quality, and language. For example, Canada and the United States use 10-digit HTS codes, whereas Viet Nam and other ASEAN members use 8-digit AHTN codes \parencite{VietnamMOF2022Circular31}. Rate-line structures also differ: Canada relies on digits 7--8 of a 10-digit code, while Viet Nam uses digits 7--8 within an 8-digit code \parencite{CBSA2026Tariff,VietnamMOF2022Circular31,UPS2025}. Accordingly, the framework’s output structure must be adapted to each jurisdiction’s tariff architecture.

The RAG component must also incorporate country-specific legal instruments. While Section and Chapter Notes are shared, national rules and supplementary notes differ. For instance, Canada applies Canadian Rules, whereas ASEAN uses AHTN-specific Supplementary Explanatory Notes \parencite{CBSAGuideCanadianRules,VietnamCustoms2023SEN}. Retrieval quality further depends on the availability of official, machine-readable tariff documentation, which varies widely across WCO members and may constrain performance in less digitized jurisdictions.

Transferability is also limited by data, expertise, legal complexity, and language. Comparable labeled datasets to the 3,300 domain-expert-labeled records used in this experiment may be difficult to obtain, as they require substantial classification expertise and resources. Jurisdictions with more complex supplementary notes, exclusions, or interpretive rules may require additional adaptation. The human-in-the-loop mechanism likewise depends on experienced local classification experts, whose availability cannot be assumed. Finally, because current models are predominantly trained in English, non-English jurisdictions may need to translate official tariff materials. Such translation can increase costs, reduce retrieval quality, and compromise legal precision, especially where terminology lacks direct equivalents or translated texts lack legal authority. Given the path-dependent nature of HTS classification, errors at higher hierarchical levels can substantially reduce final code accuracy.

\subsection{Framework Impact to Maritime Logistics and Smart Ports}
Maritime trade is the backbone of global trade, accounting for a dominant share of the transportation of traded goods \parencite{WorldBank2025ShippingPorts}. According to  \textcite{WorldBank2025ShippingPorts}, 80\% of the world’s traded goods are transported by sea. The volume of maritime trade reached 12,293 million tons in 2023 \parencite{UNCTAD2024RMT}. Within this sector, developing countries are major participants, accounting for 55\% of seaborne exports and 61\% of imports \parencite{WorldBank2025ShippingPorts}.
Traditionally, the HTS classification process in maritime logistics has been manual \parencite{yuvraj2025atlas,JudyEtAl2024BenchmarkingHTS}. Since maritime ports must process enormous volumes of diverse shipments under narrow clearance windows, this manual approach is labor-intensive, error-prone, and increasingly inadequate at port scale \parencite{JudyEtAl2024BenchmarkingHTS}.  As established in Section 2.1, any misclassification of HTS codes carries regulatory and financial consequences, including shipment delays, penalties, and supply chain disruptions. Currently, AI and digital platforms are increasingly incorporated into smart ports in order to improve the efficiency of port operations and maritime logistics \parencite{TeixeiraEtAl2026SmartGates, LiuYuen2025AISeaports, UNCTAD2024RMT}. However, an automated HTS classification system leveraging multi-agentic LLMs has received limited attention in both research and industry practice \parencite{TeixeiraEtAl2026SmartGates, LiuYuen2025AISeaports}.

As smart ports increasingly integrate AI, IoT, and digital platforms \parencite{TeixeiraEtAl2026SmartGates, LiuYuen2025AISeaports, UNCTAD2024RMT}, automated HTS classification naturally fits into this system because it can improve the quality of data inputs for cargo management systems and enhance regulatory compliance across the entire maritime and port operations sectors, including customs officers, importers, and operators. The human-in-the-loop deliberation and the human-centered application of the proposed framework align it with the governance principles of smart ports, restricting technology to augment rather than entirely replace human roles in the decision-making process \parencite{TeixeiraEtAl2026SmartGates,DagistanEtAl2025SmartPorts}. The decreasing reliance on manual performance, combined with increasing automation, creates substantial opportunities to enhance the system's operational efficiency while minimizing operational costs and resource use. Most importantly, this balanced human-AI collaboration upholds the industry's integrity and accountability, contributing to a more resilient, technology-enabled maritime future.

\section{Conclusion}
This paper presents an agentic LLM framework for Canadian 10-digit HTS code classification in maritime logistics and smart-port environments. By combining multi-agent information retrieval, semantic retrieval over official tariff documents, consensus-based validation, confidence estimation, and human-in-the-loop escalation, the proposed framework treats HTS classification as an evidence-based structured prediction task rather than a flat-label prediction problem. The experimental results show that exact 10-digit classification remains challenging even for advanced LLMs, especially when product descriptions are incomplete or legally decisive attributes are missing. Nevertheless, the framework improves interpretability and accountability by grounding predictions in retrieved evidence, localizing uncertainty across hierarchical code components, and escalating ambiguous cases for human clarification. Overall, this work demonstrates the potential of agentic LLMs to support more reliable, transparent, and compliance-oriented customs classification workflows in maritime logistics.

\section{Acknowledgments}
This work was supported by NSERC Discovery Grant No RGPIN2025-04478 and NSERC Discovery Supplement Award No DGECR2025-00129.
\printbibliography


\end{document}